\documentclass[sigconf]{acmart}
\AtBeginDocument{%
}

\usepackage{amsmath,amsfonts,bm}

\def\eqref#1{equation~\ref{#1}}

\def\1{\bm{1}}

\def\mE{{\mathbf{E}}}

\def\mI{{\mathbf{I}}}

\def\mX{{\mathbf{X}}}

\DeclareMathAlphabet{\mathsfit}{\encodingdefault}{\sfdefault}{m}{sl}
\SetMathAlphabet{\mathsfit}{bold}{\encodingdefault}{\sfdefault}{bx}{n}

\def\gE{{\mathcal{E}}}

\def\gG{{\mathcal{G}}}

\def\gT{{\mathcal{T}}}

\def\gV{{\mathcal{V}}}

\newcommand{\E}{\mathbb{E}}

\usepackage{booktabs}
\usepackage{multirow}
\usepackage{tabularx}
\usepackage{array}
\usepackage{caption}
\usepackage{xcolor}
\usepackage{soul}
\usepackage{colortbl}
\usepackage{algorithm}
\usepackage{algpseudocode}

\usepackage{amsmath,amssymb}

\setcopyright{acmlicensed}

\copyrightyear{2026}
\acmYear{2026}
\setcopyright{cc}
\setcctype{by}
\acmConference[WWW '26]{Proceedings of the ACM Web Conference 2026}{April 13--17, 2026}{Dubai, United Arab Emirates}
\acmBooktitle{Proceedings of the ACM Web Conference 2026 (WWW '26), April 13--17, 2026, Dubai, United Arab Emirates}
\acmPrice{}
\acmDOI{10.1145/3774904.3792489}
\acmISBN{979-8-4007-2307-0/2026/04}

\begin{document}
\title[Controllable Graph Generation via Inference-Time Tree Search Guidance]{Controllable Graph Generation with Diffusion Models \\via Inference-Time Tree Search Guidance}



\author{Jiachi Zhao}
\authornote{Equal Contribution.}
\authornote{Visiting Student at the University of Notre Dame.}
\affiliation{%
  \institution{University of Notre Dame}
  \city{Notre Dame}
  \state{Indiana}
  \country{USA}}
\email{jiachizhao2@gmail.com}

\author{Zehong Wang}
\authornotemark[1]
\authornote{Corresponding Authors.}
\affiliation{%
  \institution{University of Notre Dame}
  \city{Notre Dame}
  \state{Indiana}
  \country{USA}}
\email{zwang43@nd.edu}

\author{Yamei Liao}
\affiliation{%
  \institution{Independent Researcher}
  \city{New York City}
  \state{}
  \country{USA}}
\email{yameiliao.dt@gmail.com}

\author{Chuxu Zhang}
\affiliation{%
  \institution{University of Connecticut}
  \city{Storrs}
  \state{Connecticut}
  \country{USA}}
\email{chuxu.zhang@uconn.edu}

\author{Yanfang Ye}
\authornotemark[3]
\affiliation{%
  \institution{University of Notre Dame}
  \city{Notre Dame}
  \state{Indiana}
  \country{USA}}
\email{yye7@nd.edu}

\renewcommand{\shortauthors}{Jiachi Zhao, Zehong Wang, Yamei Liao, Chuxu Zhang, and Yanfang Ye}

\definecolor{Gray}{gray}{0.95}
\definecolor{Blue1}{RGB}{136, 190, 220}
\definecolor{Blue2}{RGB}{218, 232, 245}
\definecolor{Blue3}{RGB}{239, 248, 253}
\definecolor{darkgreen}{RGB}{0, 100, 0}
\definecolor{darkred}{RGB}{139, 0, 0}


\begin{abstract}
    Graph generation is a fundamental problem in graph learning with broad applications across Web-scale systems, knowledge graphs, and scientific domains such as drug and material discovery. Recent approaches leverage diffusion models for step-by-step generation, yet unconditional diffusion offers little control over desired properties, often leading to unstable quality and difficulty in incorporating new objectives. Inference-time guidance methods mitigate these issues by adjusting the sampling process without retraining, but they remain inherently local, heuristic, and limited in controllability. To overcome these limitations, we propose TreeDiff, a Monte Carlo Tree Search (MCTS) guided dual-space diffusion framework for controllable graph generation. TreeDiff is a plug-and-play inference-time method that expands the search space while keeping computation tractable. Specifically, TreeDiff introduces three key designs to make it practical and scalable: (1) a macro-step expansion strategy that groups multiple denoising updates into a single transition, reducing tree depth and enabling long-horizon exploration; (2) a dual-space denoising mechanism that couples efficient latent-space denoising with lightweight discrete correction in graph space, ensuring both scalability and structural fidelity; and (3) a dual-space verifier that predicts long-term rewards from partially denoised graphs, enabling early value estimation and removing the need for full rollouts. Extensive experiments on 2D and 3D molecular generation benchmarks, under both unconditional and conditional settings, demonstrate that TreeDiff achieves state-of-the-art performance. Notably, TreeDiff exhibits favorable inference-time scaling: it continues to improve with additional computation, while existing inference-time methods plateau early under limited resources.
\end{abstract}



\maketitle

\section{Introduction}

\begin{figure}[!t]
    \centering
    \includegraphics[width=\linewidth]{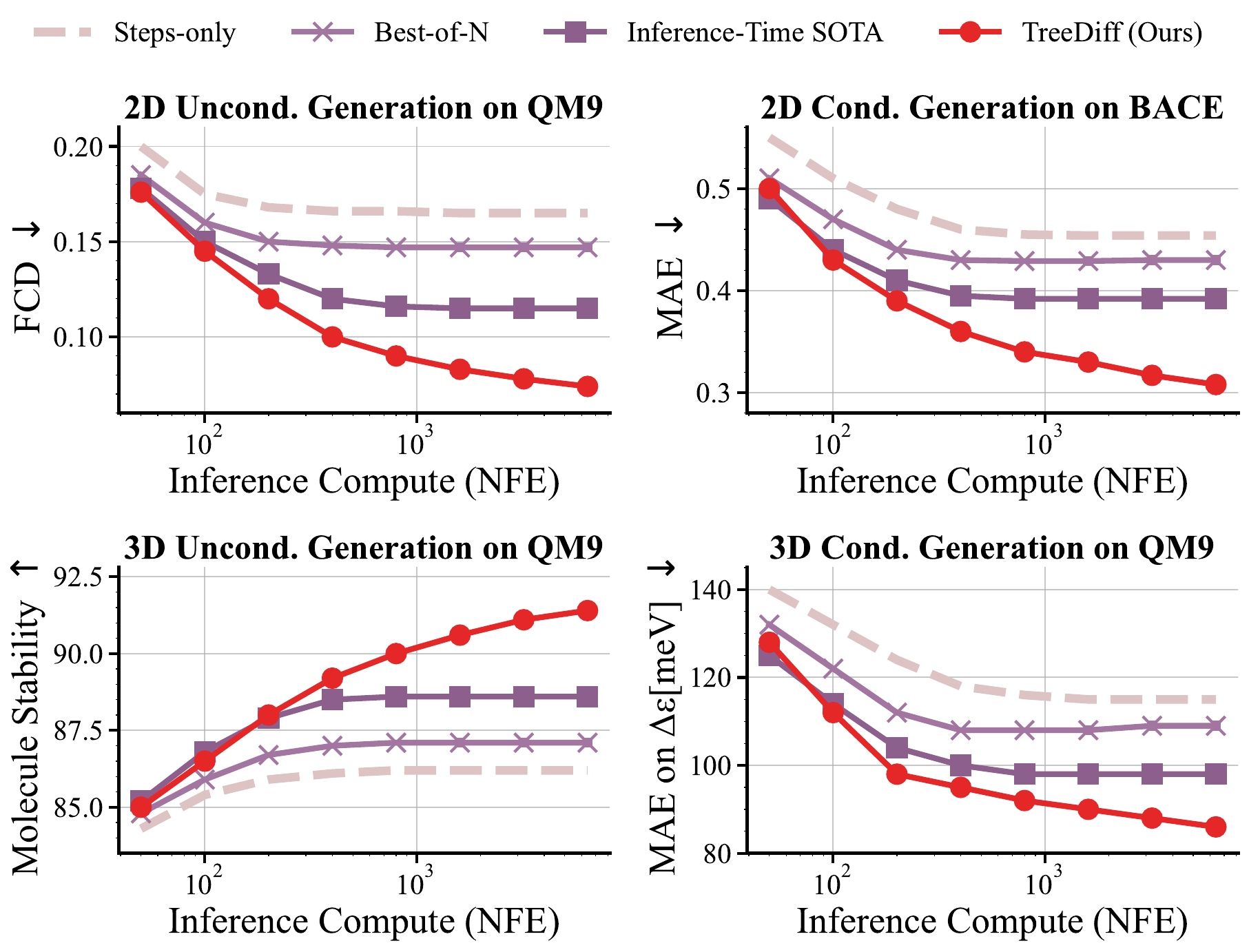}
    \vspace{-20pt}
    \caption{Inference-time scaling behavior on four generation benchmarks. TreeDiff exhibits consistent gains with additional inference computation, whereas existing approaches—including standard diffusion \citep{jo2022score}, Best-of-N \citep{ma2025inference}, and state-of-the-art inference-time guidance \citep{li2024derivative}—saturate quickly, highlighting the scalability of our approach.}
    \label{fig:scaling behavior}
    \vspace{-10pt}
\end{figure}

Graphs are powerful structures for modeling complex relationships among entities~\citep{kipf2017semisupervised,hamilton2017inductive,velicković2018graph,wang2024gft,wang2025scalable} and form the backbone of numerous Web-related systems~\citep{zhao2021multi,ju2022grape}, such as knowledge graphs, recommender networks, and online interaction graphs. The ability to generate and manipulate such graphs under domain-specific constraints has become a fundamental problem in graph learning~\citep{bonifati2020graph,simonovsky2018graphvae,shi2020graphaf,vignac2023digress,liu2024graph}. For example, in molecular and material design~\citep{liu2024graph}—where molecules can be viewed as structured graphs—generation models aim to discover new candidates that satisfy specific chemical or physical properties. Similarly, in broader Web contexts, graph generation underpins applications such as link prediction~\citep{hamilton2017inductive}, knowledge graph completion~\citep{chen2020knowledge}, and the simulation of evolving information networks~\citep{myers2014information}. Despite these diverse applications, generating graphs that are valid, diverse, and aligned with complex objectives remains a significant algorithmic challenge.

\begin{figure}[!t]
    \centering
    \includegraphics[width=\linewidth]{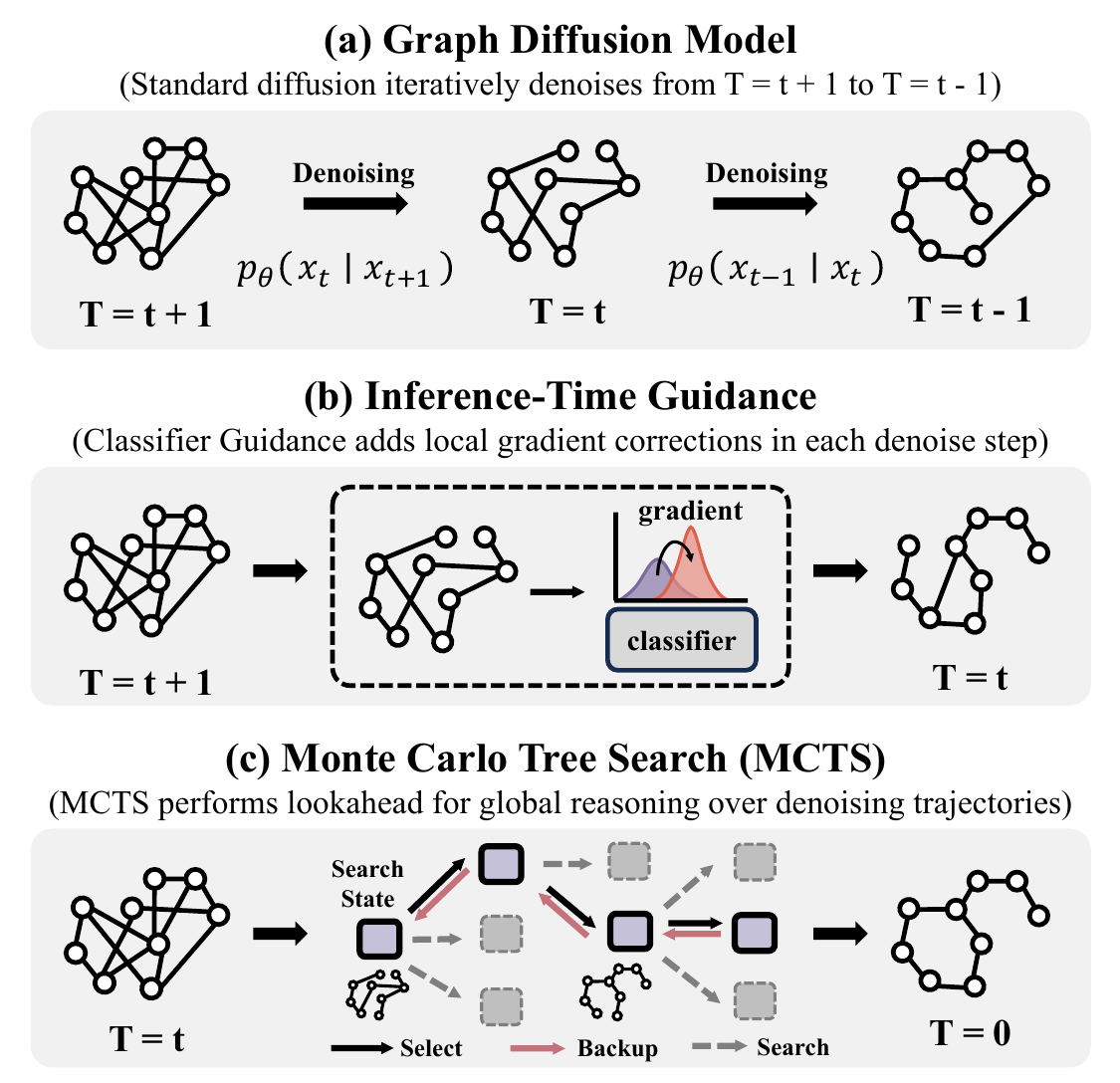}
    \caption{Comparison among diffusion-based graph generation approaches. (a) Standard graph diffusion iteratively denoises each step without correction. (b) Inference-time guidance introduces local corrections via classifier gradients but remains short-sighted. (c) Monte Carlo Tree Search (MCTS) enables structured lookahead over denoising trajectories, allowing for globally balanced exploration and exploitation.}
    \label{fig:motivation}
\end{figure}

Various generative models have been explored for graph generation~\citep{jensen2019graph,brown2019guacamol,jin2018junction,shi2020graphaf,zang2020moflow}, among which diffusion models~\citep{vignac2023digress,jo2022score,hoogeboom2022equivariant,xu2023geometric,liu2024graph} have recently gained prominence due to their ability to generate diverse, high-quality samples. Specifically, a diffusion model~\citep{ho2020denoising} learns the reverse denoising process during training to approximate the data distribution. During inference, it starts from noise and iteratively denoises to recover valid samples, ensuring consistency between training and inference. Despite the success of diffusion models in graph generation, they face two major limitations that hinder their practicality. First, controllability is limited~\citep{peebles2023scalable,podell2024sdxl}: when trained without explicit conditions or constraints, diffusion models primarily approximate the unconditional data distribution, lacking a principled way to enforce target properties during sampling. Second, generation can be unstable~\citep{watson2023novo}: the stochasticity of noise sampling often leads to invalid or low-quality outputs, reflecting the inherent fragility of the denoising process. These issues are especially severe in graph generation, where tasks like drug discovery demand molecules with domain-specific attributes (e.g., high QED, low SAS), yet such properties are rarely optimized during training. Moreover, graphs are far less tolerant to perturbations than images—even minor structural changes can drastically alter graph semantics~\citep{wang2024tackling,wang2025git}.

To this end, recent works have proposed inference-time guidance methods~\citep{ma2025inference,dorna2024tagmol,li2024derivative}, which steer the sampling process at test time without altering training. Specifically, these methods refine sampling under given conditions and prune or adjust invalid steps when necessary, thereby enabling more stable and controllable generation. For instance, \citet{ma2025inference} maintain a pool of candidate noises and perform zero-order search to expand promising branches during inference for better generation quality. Similarly, \citet{dorna2024tagmol} leverage an auxiliary classifier as a reward model to estimate the quality of intermediate states and inject gradient signals for controllable graph generation. Despite their effectiveness, existing approaches remain fundamentally local: they make decisions based on step-level feedback, greedily correcting or discarding samples according to the current state without anticipating long-term consequences. This reliance on local information limits their ability to plan across long denoising trajectories or to satisfy complex, multi-objective constraints, leaving controllability and stability still restricted~\citep{yang2023diffusion}. Empirically, this limitation manifests in poor inference-time scaling: as illustrated in Figure \ref{fig:scaling behavior}, increasing computation does not improve—and can even harm—the performance of existing approaches, underscoring their inability to benefit from additional computing resources.

To overcome the limitation, we propose using Monte Carlo Tree Search (MCTS)~\citep{silver2016mastering} as a novel inference-time controller for graph generation. Unlike greedy stepwise heuristics, MCTS enables lookahead search, evaluates future trajectories, and integrates multiple objectives in a principled manner, offering a more stable and controllable generation process. However, directly applying MCTS to diffusion-based graph generation is non-trivial and introduces several unique challenges. (1) \textit{Tree Depth Explosion}: Diffusion models typically require hundreds or even thousands of denoising steps, making one-step-per-node expansion computationally infeasible. This calls for mechanisms that can reduce the effective depth of the search while preserving the overall denoising trajectory. (2) \textit{Ill-Defined Node Expansion Space}: Defining the expansion space is also problematic. Denoising directly in graph space ensures stability and validity but is extremely slow~\citep{vignac2023digress}, while relying solely on embedding space is efficient but unstable~\citep{dorna2024tagmol}, as small perturbations can lead to drastic structural changes. This highlights the need for strategies that balance efficiency with structural reliability during node expansion. (3) \textit{Costly and Complex Evaluation}: MCTS requires reliable evaluation of future outcomes, yet in graph diffusion, full rollouts are prohibitively expensive: one must complete the entire denoising trajectory and compute molecular properties (e.g., QED, SAS), while also enforcing validity constraints. This makes it essential to design evaluation schemes that approximate long-term quality efficiently while pruning infeasible branches early.

To overcome these challenges, we introduce \textbf{TreeDiff}, a MCTS-guided dual-space diffusion framework for controllable graph generation.
Specifically, TreeDiff incorporates three key designs: (1) a macro-step expansion strategy that reduces tree depth by grouping multiple denoising steps into one node, (2) a dual-space denoising mechanism that combines efficiency in the embedding space with stability in the graph space, and (3) a dual-space verifier that provides long-term value estimates to guide global search under complex validity constraints. Extensive experiments on 2D and 3D molecular benchmarks show that TreeDiff achieves state-of-the-art performance on both unconditional and conditional generation tasks. Notably, as shown in Figure \ref{fig:scaling behavior}, TreeDiff exhibits favorable inference-time scaling behavior—continuing to improve as more computation is allocated, whereas existing methods quickly plateau.

\section{Related Work}

\noindent\textbf{Graph Generation.} 
Graph generation~\citep{bonifati2020graph,thompson2022evaluation,tang2022graph} learns distributions over graph topology and attributes, enabling both unconditional and conditional synthesis. However, validity, property control, and efficiency remain difficult to achieve simultaneously~\citep{maziarz2021learning,xu2021learning}. Prior work mainly falls into four families: autoregressive models allow rule injection but are order-sensitive and slow~\citep{you2018graphrnn,liao2019efficient,shi2020graphaf}; VAE-based methods support latent interpolation but impose size limits~\citep{simonovsky2018graphvae,jin2018junction}; GAN-based approaches provide one-shot fidelity but suffer from instability~\citep{de2018molgan}; and flow-based models yield tractable likelihoods but face dequantization bias~\citep{zang2020moflow,luo2021graphdf}. More recently, diffusion models~\citep{sohl2015deep,ho2020denoising,song2020score} have emerged as the dominant paradigm, offering high-quality graph generation.

\noindent\textbf{Graph Diffusion Models.} 
Diffusion models~\citep{ho2020denoising,rombach2022high,gong2022diffuseq,nie2025large} have recently been adapted to graphs by injecting noise into discrete topology and attributes and learning a reverse denoising process. Representative works include GDSS~\citep{jo2022score}, which learns joint node \& adjacency scores in a continuous-time SDE; DiGress~\citep{vignac2022digress}, which applies discrete transformations over nodes and edges; GeoDiff~\citep{xu2022geodiff}, which performs SE(3)-equivariant diffusion for 3D conformations; and NVDiff~\citep{chen2022nvdiff}, which denoises node embeddings for scalability. Extensions further explore conditional generation~\citep{liu2023data,jensen2019graph,xie2021mars,lee2023exploring,guan20233d,guan2024decompdiff} and efficient sampling~\citep{luo2023fast,cho2023multi}. Despite these advances, controllability and stability remain open challenges: enforcing graph-level constraints consistently across multi-step denoising is still unreliable. Our proposed TreeDiff directly addresses these issues at inference time without retraining.

\noindent\textbf{Inference-Time Guidance.} 
For graph diffusion, inference-time guidance~\citep{uehara2025inference} offers a way to enforce validity and target properties without retraining, by steering the denoising trajectory at test time. Existing approaches can be grouped into three categories: (i) classifier guidance~\citep{dhariwal2021diffusion,ho2022classifier,kim2022diffusionclip,chung2022diffusion,dorna2024tagmol}, which trains auxiliary models to score graph properties (e.g., QED/SA) and provide gradients or rewards; (ii) sampling-based methods~\citep{wu2023practical,cardoso2023monte,phillips2024particle,li2024derivative,ma2025inference,nichol2021glide}, which resample from candidate noises or trajectories and retain the best ones; and (iii) search-based methods~\citep{yoon2025monte,jain2025diffusion,oshima2025inference,li2024derivative,hubert2021learning,feng2023alphazero,hao2023reasoning,kajita2020autonomous,swanson2024generative}, which keep a small frontier and apply heuristics or light backtracking. While these methods improve controllability, in graph domains they remain limited: feedback is often noisy, validity is fragile, and scaling candidates rarely yields consistent gains. Our proposed TreeDiff instead employs MCTS for global planning, integrating property guidance with validity checks along full trajectories to achieve robust, controllable, and scalable inference.
\section{Preliminary}

\noindent\textbf{Problem Definition.}
We study the problem of graph generation. Formally, let $\gG = (\gV, \gE, \mX, \mE)$ denote a graph, where $\gV$ is the set of nodes with $|\gV| = N$, $\gE \subseteq \gV \times \gV$ is the set of edges with $|\gE| = E$, and $\mX$ and $\mE$ represent the node and edge feature matrices, respectively. Each node $v \in \gV$ is associated with a feature vector $\mathbf{x}_v \in \mathbb{R}^{d_n}$, and each edge $e \in \gE$ is associated with a feature vector $\mathbf{e}_e \in \mathbb{R}^{d_e}$ (if edge attributes are available). Given a target distribution $\mathcal{P}_{\text{data}}$ over graphs, the goal is to learn a generative model $p_\theta(\gG)$ such that $p_\theta(\gG) \approx \mathcal{P}_{\text{data}}$. That is, the model should generate graphs that are both structurally valid and semantically consistent with the observed data, while preserving diversity and faithfully capturing the underlying distribution.

\begin{figure*}[!t]
    \centering
    \includegraphics[width=\linewidth]{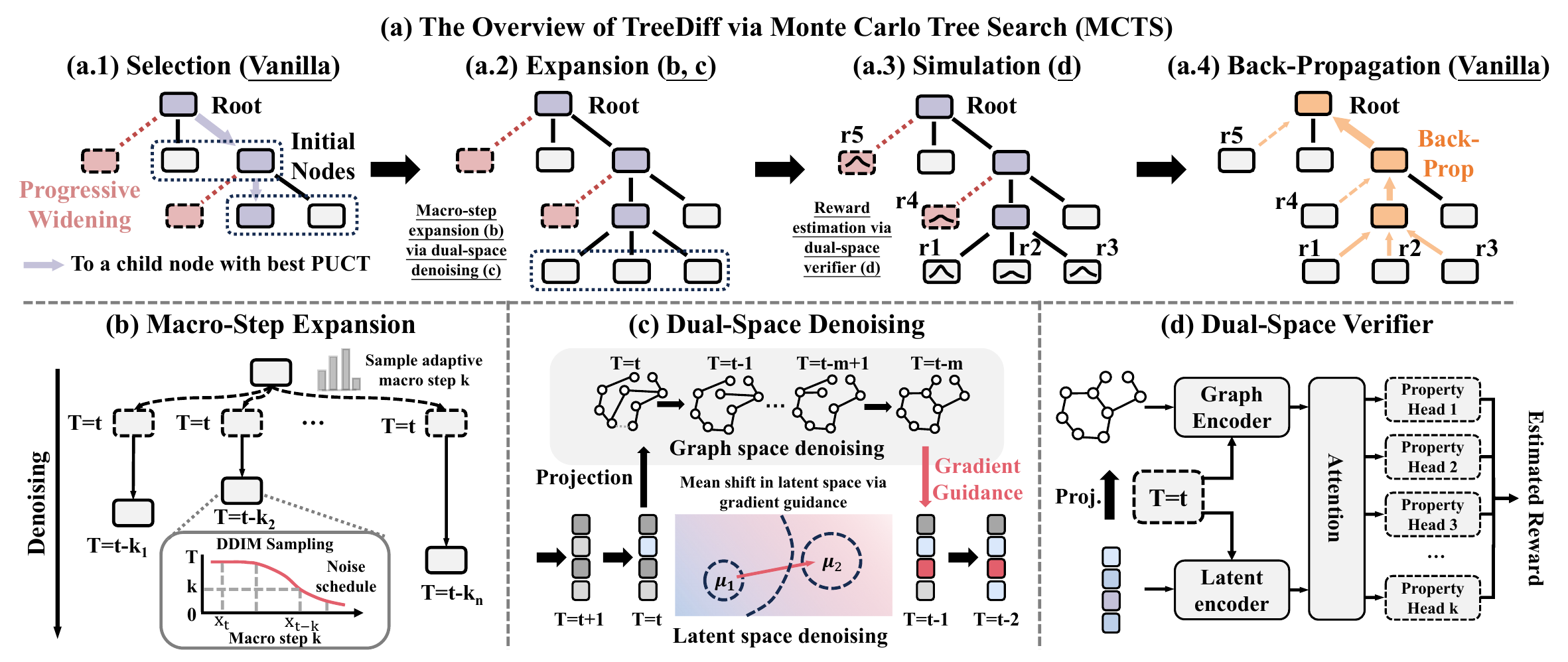}
    \caption{
        TreeDiff integrates MCTS with diffusion-based graph generation.
        (a) TreeDiff follows the standard MCTS pipeline but introduces new designs in the \textit{Expansion} and \textit{Simulation} phases.
        (b) \textit{Macro-step expansion} reduces tree depth by grouping multiple denoising updates.
        (c) \textit{Dual-space denoising} couples latent diffusion with discrete structural correction to ensure validity.
        (d) \textit{Dual-space verifier} predicts long-term rewards from latent and graph embeddings, enabling rollout-free evaluation.
    }
    \label{fig:framework}
\end{figure*}

\noindent\textbf{Graph Diffusion Model.}
We adopt the diffusion framework in the context of graph generation. Given a graph $\gG = (\gV, \gE, \mX, \mE)$, the forward process constructs a sequence of latent variables $\{ \gG_t \}_{t=0}^T$, where $\gG_0 = \gG$ and $\gG_T$ approximates a noise distribution. At each step, noise is added according to a predefined Markov transition:
\begin{equation}
    q(\gG_t \mid \gG_{t-1}) = \mathcal{N}\!\left( \sqrt{1-\beta_t}\,\gG_{t-1}, \, \beta_t \mI \right),
\end{equation}
where $\{ \beta_t \}_{t=1}^T$ is a variance schedule. The reverse process aims to learn a parameterized distribution $p_\theta(\gG_{t-1} \mid \gG_t)$ that approximates the true posterior $q(\gG_{t-1} \mid \gG_t, \gG_0)$. Sampling proceeds by iteratively denoising from Gaussian noise toward the target graph $\mathcal{G}_0$. The training objective is typically a reweighted denoising score matching loss:
\begin{equation}
    \mathcal{L}_{\text{diff}}(\theta) = \E_{t, \gG_0, \epsilon}\!\left[ \| \epsilon - \epsilon_\theta(\gG_t, t) \|^2 \right],
\end{equation}
where $\epsilon$ is Gaussian noise and $\epsilon_\theta$ is a neural network that predicts the noise component. In graph generation, diffusion can be applied either directly in the graph space (adjacency and feature matrices) or in a continuous latent space learned by an encoder.


\noindent\textbf{Monte Carlo Tree Search.}
The denoising process in diffusion models can be naturally viewed as a sequential decision problem, where each step refines the current state and local errors may accumulate into invalid or low-quality graphs.
This motivates the use of search-based methods that can plan across long denoising trajectories. Among them, MCTS is a widely used algorithm for decision-making in large spaces under uncertainty. Formally, MCTS builds a search tree $\gT$, where each node $s \in \gT$ represents a state (e.g., a partially denoised graph), and edges correspond to actions $a \in \mathcal{A}(s)$ (e.g., candidate denoising steps). Each iteration of MCTS consists of four phases: (i) \textit{Selection}, choosing a path from the root to a leaf according to an exploration–exploitation policy (e.g., choosing a partial denoising trajectory); (ii) \textit{Expansion}, adding a new child node to represent an unexplored state (e.g., the next denoising step); (iii) \textit{Simulation}, estimating the long-term outcome by rolling out a trajectory from the new node (e.g., sampling a complete graph), yielding a reward $R$; and (iv) \textit{Backpropagation},  propagating $R$ upward to update the statistics of nodes along the path. By adaptively allocating computation to promising branches, MCTS provides lookahead planning, maintains multiple candidate trajectories, and integrates domain-specific objectives.

\section{Methodology}

We propose TreeDiff, an MCTS-guided diffusion framework for controllable and stable graph generation.
TreeDiff formulates diffusion sampling as a sequential decision process and applies structured tree search to optimize denoising trajectories at inference time.
Directly combining MCTS with diffusion is challenging due to three factors: (i) extremely long denoising horizons, (ii) the efficiency--stability trade-off between latent and graph spaces, and (iii) the high cost of full-trajectory evaluation.
To overcome these challenges, TreeDiff introduces three components (Figure~\ref{fig:framework}):
(1) \emph{macro-step expansion} to compress long trajectories into coarse transitions,
(2) \emph{dual-space denoising} to couple efficient latent updates with graph-level structural correction, and
(3) a \emph{dual-space verifier} to estimate long-term rewards without full rollouts.
These designs enable adaptive inference-time computation and substantially improve generation controllability.
Implementation details are provided in Appendix~\ref{sec:imp}.

\subsection{Macro-Step Expansion}

In MCTS, the \emph{Expansion} phase determines how new nodes are added to the search tree.
For diffusion-based generation, each reverse step only slightly refines the graph state, resulting in extremely deep trajectories.
Expanding at every step would yield hundreds or thousands of near-duplicate layers, making exploration inefficient and long-horizon planning impractical.
We redefine the expansion granularity using a \emph{macro-step expansion} strategy that groups multiple denoising steps into a single transition.
Instead of expanding from $\gG_t$ to $\gG_{t-1}$, TreeDiff directly expands to $\gG_{t-k}$ for a horizon $k$, substantially reducing tree depth while preserving overall denoising dynamics.
This converts expansion from step-wise local moves into hierarchical exploration over the denoising trajectory.

Formally, let $\gG_t$ denote the intermediate graph at reverse step $t$.
A single update is written as
$\gG_{t-1}=\Phi_\theta(\gG_t,t;\xi_t)$,
where $\Phi_\theta$ is the learned reverse transition and $\xi_t\!\sim\!\mathcal{N}(0,\mI)$.
A $k$-step macro transition is defined by composition:
\begin{align}
    \nonumber
    \gG_{t-k} & = \Phi_\theta^{(k)}(\gG_t,t;\xi_{t-k+1:t})                                                        \\
              & \triangleq \Phi_\theta\!\big(\Phi_\theta^{(k-1)}(\gG_t,t;\xi_{t-k+2:t}),\,t-k+1;\xi_{t-k+1}\big),
\end{align}
with $\xi_{t-k+1:t}=\{\xi_{t-k+1},\ldots,\xi_t\}$.
During \textit{Expansion}, TreeDiff samples a small set of candidate macro-steps and inserts the resulting states as children of $\gG_t$ in the search tree $\gT$.

\noindent\textbf{Adaptive Step Sampling.}
Rather than fixing $k$, TreeDiff adapts the adaptive expansion.
Given remaining diffusion steps $T_{\text{rem}}=t$ and remaining tree depth $D_{\text{rem}}$, we set
$k_{\text{base}}=\frac{T_{\text{rem}}}{D_{\text{rem}}}$,
and sample
\begin{equation}
    k \sim \mathcal{N}\!\big(k_{\text{base}},(\sigma_k k_{\text{base}})^2\big), \quad 1\le k\le T_{\text{rem}},
\end{equation}
where $\sigma_k=0.1$.
This stochastic scheduling enables both coarse long-range moves and fine local refinement, improving the exploration--exploitation trade-off.
When $T_{\text{rem}}<k_{\text{base}}$, the process directly denoises to $\gG_0$.

\subsection{Dual-Space Denoising}

Macro-step expansion enables efficient exploration but raises a key question for MCTS: \emph{which representation should be used for selection and expansion?}
Each node can be represented in either latent or graph space, yet both have limitations.
Latent-space search provides smooth transitions for UCB~\citep{auer2002finite} or PUCT~\citep{silver2016mastering}, but decoded trajectories may become invalid.
Graph-space search preserves structural validity and enables property evaluation, but introduces combinatorial branching and discontinuous transitions, leading to unstable selection.
This mismatch hinders effective long-horizon planning.
To address it, TreeDiff adopts \emph{dual-space denoising}, coupling continuous latent diffusion with lightweight graph-space correction.

Specifically, dual-space denoising governs each $k$-step expansion by alternating continuous diffusion and discrete refinement.
Let $\gG_t^{\text{latent}}$ and $\gG_t^{\text{structure}}$ denote the latent and decoded graph states at step $t$.
TreeDiff first performs $n$ latent denoising steps to obtain $\gG_{t-n}^{\text{latent}}$, then decodes it as
$\gG_{t-n}^{\text{structure}}=\mathrm{Dec}(\gG_{t-n}^{\text{latent}},t-n)$.
A lightweight denoiser $p_\psi$ applies $m\!\ll\!n$ categorical refinement steps to enforce topological consistency, producing $\gG_{t-n-m}^{\text{structure}}$, which is re-encoded as
$\gG_{t-n-m}^{\text{latent}}=\mathrm{Enc}(\gG_{t-n-m}^{\text{structure}},t-n-m)$.

Rather than treating this reprojected latent as a new trajectory, TreeDiff interprets it as a directional guide that reshapes the remaining diffusion.
A guidance vector is computed as
\begin{equation}
    g_{t-n} = \nabla_{\gG_{t-n}^{\text{latent}}} \!\left(-\frac{1}{2\sigma_g^2} \big\|\gG_{t-n-m}^{\text{latent}} - \gG_{t-n}^{\text{latent}} \big\|_2^2\right)
    = \frac{\gG_{t-n-m}^{\text{latent}} - \gG_{t-n}^{\text{latent}}}{\sigma_g^2},
\end{equation}
which acts as the gradient of an implicit potential attracting the trajectory toward structure-consistent regions.
For each subsequent reverse step $t' < k-n$, the guided posterior is defined as
\begin{equation}
    p_\theta(\gG_{t'-1}^{\text{latent}} \mid \gG_{t'}^{\text{latent}})
    =
    \mathcal{N}\!\Big(
    \tilde{\mu}_\theta(\gG_{t'}^{\text{latent}}, t')
    + h_{t'}\,\tilde{\beta}_{t'}\,g_{t-n},\;
    \tilde{\beta}_{t'}\,\mI
    \Big),
\end{equation}
where $\tilde{\mu}_\theta$ is the standard posterior mean, $\tilde{\beta}_{t'}$ is the variance schedule, and $h_{t'}$ modulates the strength of structural guidance. Intuitively, the first $n$ latent steps explore freely under unconstrained diffusion, while the subsequent guided updates are softly attracted toward structurally valid manifolds. This allows MCTS to conduct selection and value backup entirely within the latent domain, while periodically anchoring trajectories in the discrete space to enforce validity and stability across long rollouts.

\noindent\textbf{Trajectory Distillation for Auxiliary Modules.}
Inspired by the success of large language models in distilling high-quality reasoning trajectories, we adopt a similar principle to train the auxiliary components of TreeDiff—namely, the time-conditioned encoder $\mathrm{Enc}(\cdot)$, decoder $\mathrm{Dec}(\cdot)$, and discrete denoiser $p_\psi(\cdot)$. Rather than training them jointly with diffusion, we distill knowledge from pre-collected generation trajectories so that these modules capture diffusion-consistent dynamics. Concretely, we generate 500 full denoising trajectories $\{\gG_t\}_{t=0}^T$ using a pretrained diffusion model and treat them as teacher demonstrations. The encoder-decoder pair is trained as a time-conditioned variational autoencoder that reconstructs intermediate graphs along these trajectories:
\begin{align}
    \nonumber
    \mathcal{L}_{\text{VAE}}
     & = \mathbb{E}_{\gG_t}\!\left[\|\gG_t - \mathrm{Dec}(\mathrm{Enc}(\gG_t,t),t)\|_2^2\right] \\
     & \quad + \lambda_{\text{KL}}\,
    \mathrm{KL}\!\big(q(z_t|\gG_t,t)\,\|\,\mathcal{N}(0,\mI)\big),
\end{align}
which encourages temporally smooth and semantically aligned latent representations.
Meanwhile, the graph-space denoiser $p_\psi$ is trained on consecutive teacher states $(\gG_{t+1}\!\rightarrow\!\gG_t)$ to predict the next denoised structure:
\begin{equation}
    \mathcal{L}_{\text{denoise}}
    = \mathbb{E}_{t}\!\left[
        \|\hat{\mX}_t - \mX_t\|_2^2
        + \|\hat{\mE}_t - \mE_t\|_2^2
        \right],
    (\hat{\mX}_t,\hat{\mE}_t)=p_\psi(\gG_{t+1},t).
\end{equation}
The architectures of these modules are implementation-agnostic.

\begin{table*}[!t]
    \caption{Conditional graph generation results on 2D molecules across diversity (Div., \%), similarity (Sim., \%), mean absolute error (MAE), and accuracy (Acc., \%). Boldface and underline indicate the best and second-best, and A.R. is the average ranking.}
    \label{tab:2d condition}
    \resizebox{0.95\linewidth}{!}{
        \begin{tabular}{cl cccc | cccc | cccc | c}
            \toprule                                                  &                                      & \multicolumn{4}{c}{\bf Synth. \& BACE} & \multicolumn{4}{c}{\bf Synth. \& BBBP} & \multicolumn{4}{c}{\bf Synth. \& HIV}                                                                                                                                                                                                                                                                      \\
            \cmidrule(lr){3-6}\cmidrule(lr){7-10}\cmidrule(lr){11-14} & \textbf{Method}                      & \textbf{Div. $\uparrow$}               & \textbf{Sim. $\uparrow$}               & \textbf{MAE $\downarrow$}             & \textbf{Acc. $\uparrow$} & \textbf{Div. $\uparrow$} & \textbf{Sim. $\uparrow$} & \textbf{MAE $\downarrow$} & \textbf{Acc. $\uparrow$} & \textbf{Div. $\uparrow$} & \textbf{Sim. $\uparrow$} & \textbf{MAE $\downarrow$} & \textbf{Acc. $\uparrow$} & \textbf{A.R.} \\
            \midrule \multirow{4}{*}{\rotatebox{90}{Tradition}}       & Graph-GA~\citep{jensen2019graph}     & 85.85                                  & \textbf{98.05}                         & 0.963                                 & 46.90                    & 89.50                    & {\textbf{95.09}}         & 1.208                     & 30.15                    & 89.93                    & \ul{96.61}               & 0.984                     & 60.35                    & 7.3           \\
                                                                      & MARS~\citep{xie2021mars}             & 83.38                                  & 88.27                                  & 1.012                                 & 51.84                    & 86.37                    & 76.96                    & 1.225                     & 51.89                    & 87.64                    & 65.17                    & 0.969                     & 64.55                    & 9.5           \\
                                                                      & LSTM-HC~\citep{brown2019guacamol}    & 81.46                                  & 79.82                                  & 0.921                                 & 58.16                    & 88.83                    & 89.32                    & 0.997                     & 55.90                    & 90.91                    & 91.45                    & 0.948                     & 67.36                    & 7.8           \\
                                                                      & JTVAE-BO~\citep{jin2018junction}     & 66.82                                  & 72.81                                  & 0.992                                 & 46.28                    & 74.58                    & 58.21                    & 1.162                     & 49.58                    & 80.55                    & 41.73                    & 1.236                     & 48.50                    & 11.8          \\
            \midrule \multirow{5}{*}{\rotatebox{90}{Diffusion}}       & DiGress~\citep{vignac2023digress}    & 88.62                                  & 69.42                                  & 2.068                                 & 50.61                    & 90.98                    & 68.05                    & 2.366                     & 65.36                    & \ul{91.94}               & 85.62                    & 1.922                     & 53.35                    & 9.1           \\
                                                                      & DiGress v2~\citep{vignac2023digress} & 88.12                                  & 70.27                                  & 2.337                                 & 51.13                    & 91.07                    & 63.36                    & 2.269                     & 65.31                    & 91.93                    & 84.76                    & 1.593                     & 53.31                    & 9.2           \\
                                                                      & GDSS~\citep{jo2022score}             & 87.56                                  & 27.08                                  & 1.642                                 & 50.36                    & 84.15                    & 26.72                    & 1.379                     & 50.37                    & 78.17                    & 10.32                    & 1.252                     & 48.30                    & 11.9          \\
                                                                      & MOOD~\citep{lee2023exploring}        & \ul{89.02}                             & 25.87                                  & 1.885                                 & 50.62                    & {\textbf{92.73}}         & 17.15                    & 2.028                     & 49.03                    & \textbf{92.80}           & 13.61                    & 2.314                     & 51.06                    & 9.8           \\
                                                                      & Graph DiT~\citep{liu2024graph}       & 82.38                                  & 87.52                                  & \ul{0.400}                            & 91.35                    & 88.56                    & 93.29                    & \ul{0.355}                & 94.17                    & 89.74                    & 95.75                    & \ul{0.309}                & \ul{97.77}               & 4.8           \\
            \midrule \multirow{5}{*}{\rotatebox{90}{Inference-Time}}  & Best-of-N~\citep{ma2025inference}    & 87.29                                  & 88.50                                  & 0.480                                 & 91.01                    & 89.29                    & 92.41                    & 0.420                     & 93.80                    & 88.54                    & 95.61                    & 0.360                     & 97.01                    & 5.8           \\
                                                                      & Beam Search~\citep{ma2025inference}  & 86.03                                  & \ul{89.21}                             & 0.455                                 & 91.40                    & 88.53                    & 92.86                    & 0.405                     & 94.11                    & 87.88                    & 95.84                    & 0.345                     & 97.20                    & 5.3           \\
                                                                      & TAGMol~\citep{dorna2024tagmol}       & 88.64                                  & 87.45                                  & 0.430                                 & \ul{91.60}               & 90.45                    & 91.43                    & 0.372                     & \ul{94.46}               & 89.62                    & 95.17                    & 0.325                     & 97.56                    & 4.5           \\
                                                                      & SVDD~\citep{li2024derivative}        & 86.51                                  & 89.03                                  & 0.510                                 & 90.56                    & 88.81                    & 92.69                    & 0.460                     & 93.50                    & 88.04                    & 95.73                    & 0.395                     & 96.83                    & 6.3           \\
            \cmidrule{2-15}                                           & \textbf{TreeDiff (Ours)}             & \textbf{89.30}                         & 87.82                                  & \textbf{0.392}                        & \textbf{91.83}           & \ul{91.62}               & \ul{93.63}               & \textbf{0.342}            & \textbf{94.81}           & 90.41                    & \textbf{96.90}           & \textbf{0.302}            & \textbf{97.90}           & 1.9           \\
            \bottomrule
        \end{tabular}
    }
\end{table*}

\begin{table*}[!t]
    \caption{Conditional graph generation results on 3D molecular benchmark QM9 (ID \& OOD). We report mean absolute error (MAE) on six SE(3)-invariant quantum properties: polarizability $\alpha$ (Bohr$^3$), gap $\Delta\varepsilon$ (meV), HOMO energies $\varepsilon_H$ (meV), LUMO energies $\varepsilon_L$ (meV), dipole moment $\mu$ (D), and heat capacity $C_v$ ($cal/mol \cdot K$). }
    \label{tab:3d condition}
    \centering
    \resizebox{0.95\linewidth}{!}{
        \begin{tabular}{clcccccc|cccccc|c}
            \toprule                                        &                                               & \multicolumn{6}{c}{\bf In-Distribution (MAE $\downarrow$)} & \multicolumn{6}{c}{\bf Out-Of-Distribution (MAE $\downarrow$)} &                                                                                                                                                                                                                                                    \\
            \cmidrule(lr){3-8}\cmidrule(lr){9-14}           & \textbf{Method}                               & $\alpha$(Bohr$^3$)                                         & $\Delta\varepsilon$(meV)                                       & $\varepsilon_H$(meV) & $\varepsilon_L$(meV) & $\mu$(D)      & $C_v$($\frac{cal}{mol}$K) & $\alpha$(Bohr$^3$) & $\Delta\varepsilon$(meV) & $\varepsilon_H$(meV) & $\varepsilon_L$(meV) & $\mu$(D)       & $C_v$($\frac{cal}{mol}$K) & \textbf{A.R.} \\\midrule
                                                            & Random                                        & 41.00                                                      & 193.36                                                         & 103.30               & 121.83               & 8.40          & 13.56                     & 73.03              & \ul{344.43}              & 183.00               & 217.01               & 14.96          & 24.15                     & 6.3           \\            \midrule
            \multirow{2}{*}{\rotatebox{90}{Diff.}}          & EDM\cite{wu2022diffusion}                     & 20.15                                                      & 287.00                                                         & 158.70               & 166.20               & 7.01          & 13.63                     & 55.70              & 1561.90                  & 1196.80              & 228.20               & 19.13          & 38.42                     & 7.6           \\
                                                            & LDM\textendash 3DG \cite{you2024latent}       & 15.56                                                      & 107.14                                                         & \textbf{54.62}       & 63.08                & 6.33          & 13.66                     & 32.06              & 363.13                   & 109.30               & 178.69               & 22.18          & 31.12                     & 4.9           \\\midrule
            \multirow{5}{*}{\rotatebox{90}{Inference-Time}} & Best\mbox{-}of\mbox{-}N\cite{ma2025inference} & 16.23                                                      & 109.97                                                         & 58.11                & 67.08                & 6.27          & 13.54                     & 31.07              & 359.91                   & 110.13               & 175.11               & 14.59          & 24.07                     & 4.2           \\
                                                            & Beam Search\cite{ma2025inference}             & 16.04                                                      & 108.06                                                         & 57.14                & 66.05                & 6.29          & 13.58                     & 33.47              & 345.11                   & 103.23               & \ul{171.03}          & \ul{14.31}     & \ul{23.79}                & 3.7           \\
                                                            & TAGMol\cite{dorna2024tagmol}                  & \ul{15.33}                                                 & \ul{100.19}                                                    & \ul{55.09}                & \ul{63.06}           & \ul{6.13}     & \ul{13.47}                & \ul{31.01}         & 345.08                   & \ul{102.09}          & 171.07               & 14.77          & 24.29                     & 2.7           \\
                                                            & SVDD\cite{li2024derivative}                   & 16.78                                                      & 115.11                                                         & 60.07                & 67.39                & 6.26          & 13.57                     & 34.47              & 370.13                   & 115.76               & 178.11               & 14.83          & 23.91                     & 5.2           \\ \cmidrule(l){2-15}
                                                            & \textbf{TreeDiff (Ours)}                      & \textbf{14.92}                                             & \textbf{95.21}                                                 & 61.55           & \textbf{60.19}       & \textbf{6.07} & \textbf{13.41}            & \textbf{28.55}     & \textbf{341.76}          & \textbf{99.53}       & \textbf{170.11}      & \textbf{14.24} & \textbf{23.51}            & 1.4           \\
            \bottomrule
        \end{tabular}
    }
\end{table*}

\subsection{Dual-Space Verifier}

In MCTS, the \textit{Simulation} and \textit{Backpropagation} phases estimate long-term rewards for candidate trajectories.
However, full rollouts are prohibitively expensive for diffusion with hundreds of steps and multiple objectives.
TreeDiff therefore introduces a learned \emph{dual-space verifier} $V_\phi$ to predict terminal rewards from partially denoised states.
At step $t$, it takes both the latent representation $\gG_t^{\text{latent}}$ and the decoded structure $\gG_t^{\text{structure}}$ as input and outputs $\hat R_t \approx R(\gG_0)$.
This replaces explicit rollouts during search, enabling efficient multi-objective planning while remaining consistent with dual-space denoising.

The verifier is trained as a regression model on full denoising trajectories $\{\gG_t\}_{t=0}^{T}$ generated by a pretrained diffusion model, each labeled with the terminal reward
$R(\gG_0)=\sum_k w_k f_k(\gG_0)$.
To improve robustness, we apply stochastic augmentation by perturbing latent states
$\tilde{\gG}_t^{\text{latent}}=\gG_t^{\text{latent}}+\epsilon$, $\epsilon\sim\mathcal{N}(0,\sigma_a^2\mI)$,
and decoding them into $\tilde{\gG}_t^{\text{structure}}$.
The verifier minimizes
\begin{equation}
    \mathcal{L}_{\text{verifier}}
    =
    \E_t\!\left[
        \|V_\phi(\tilde{\gG}_t^{\text{latent}}, \tilde{\gG}_t^{\text{structure}}, t)
        -
        R(\gG_0)\|_2^2
        \right].
\end{equation}

In practice, $V_\phi$ adopts a dual-branch architecture: a graph transformer encodes $\gG_t^{\text{structure}}$, an MLP processes $\gG_t^{\text{latent}}$, and cross-attention fuses both before predicting $\hat R_t$.
During MCTS, $\hat R_t$ is used as the node value for simulation and backpropagation, enabling efficient evaluation without rollouts.

The dual-space verifier completes TreeDiff's search loop by providing fast, reliable value estimates.
Together with macro-step expansion and dual-space denoising, it enables latent-space planning while remaining grounded in discrete graph semantics.
Since $R(\gG_0)$ can combine arbitrary objectives, the verifier naturally supports controllable multi-objective generation at inference time.

\section{Experiments}

\subsection{Conditional Graph Generation}

\noindent\textbf{Conditional Generation on 2D Molecular Graphs.}
We evaluate conditional graph generation on three molecular benchmarks, following~\citet{liu2024graph}: BACE (human $\beta$-secretase~1 inhibition), BBBP (blood–brain barrier permeability), and HIV (replication inhibition). Apart from molecular properties, we additionally condition on synthetic accessibility (SAS) and synthetic complexity (SCS)~\citep{ertl2009estimation,coley2018scscore}. Datasets are split into $6{:}2{:}2$ train/validation/test sets~\citep{polykovskiy2020molecular}, and we generate 10,000 molecules per task. Evaluation metrics include diversity (Div., \%), fragment-based similarity (Sim., \%), mean absolute error (MAE), and classification accuracy (Acc., \%). Each experiment is repeated five times and averaged to mitigate randomness. We compare against a broad set of baselines: traditional methods (Graph-GA~\citep{jensen2019graph}, MARS~\citep{xie2021mars}, LSTM-HC~\citep{brown2019guacamol}, JTVAE-BO~\citep{jin2018junction}); diffusion-based generators (DiGress~\citep{vignac2023digress}, DiGress v2, GDSS~\citep{jo2022score}, MOOD~\citep{lee2023exploring}, GraphDiT~\citep{liu2024graph}); and inference-time guidance strategies, including Best-of-N~\citep{ma2025inference}, Beam Search~\citep{ma2025inference}, TAGMol~\citep{dorna2024tagmol}, and SVDD~\citep{li2024derivative}. For fairness, all inference-time methods adopt GraphDiT as the backbone, with matched compute budgets. As summarized in Table~\ref{tab:2d condition}, inference-time guidance consistently improves diffusion baselines, and TreeDiff delivers the most stable and substantial gains across all benchmarks. It achieves the best average rank (A.R.\ 1.9), outperforming both diffusion and inference-only competitors. In terms of distribution quality, TreeDiff avoids the trade-offs between diversity and similarity, maintaining both at high levels (e.g., Synth.\&BACE: Div.\ 89.30 vs.\ 82.38 and Sim.\ 87.82 vs.\ 87.52 for Graph DiT). For controllability, TreeDiff achieves stronger condition matching, with lower MAE and higher Accuracy across tasks (e.g., Synth.\&HIV: MAE 0.302 and Acc.\ 97.90\%).

\begin{table*}[!t]
    \caption{Unconditional graph generation results on 2D \textbf{QM9} and \textbf{ZINC250k} regarding validity (Valid), Fr\'echet ChemNet Distance (FCD), NSPDK kernel distance (NSPDK), and scaffold similarity (Scaf.).}
    \label{tab:2d uncondition}
    \resizebox{0.8\linewidth}{!}{
        \begin{tabular}{cl cccc | cccc | c}
            \toprule
                                                            &                                                & \multicolumn{4}{c}{\bf QM9} & \multicolumn{4}{c}{\bf ZINC250k} &                                                                                                                                                                                           \\ \cmidrule(lr){3-6}\cmidrule(lr){7-10}
                                                            & \textbf{Method}                                & \textbf{Valid $\uparrow$}   & \textbf{FCD $\downarrow$}        & \textbf{NSPDK $\downarrow$} & \textbf{Scaf. $\uparrow$} & \textbf{Valid $\uparrow$} & \textbf{FCD $\downarrow$} & \textbf{NSPDK $\downarrow$} & \textbf{Scaf. $\uparrow$} & \textbf{A.R.} \\ \midrule
                                                            & Training set                                   & 100.00                      & 0.040                            & 0.0001                      & 0.972                     & 100.00                    & 0.062                     & 0.0001                      & 0.840                     & --            \\ \midrule
            \multirow{3}{*}{\rotatebox{90}{Trad.}}          & MoFlow~\cite{zang2020moflow}                   & 91.36                       & 4.467                            & 0.0169                      & 0.145                     & 63.11                     & 20.931                    & 0.0455                      & 0.013                     & 10.4          \\
                                                            & GraphAF~\cite{shi2020graphaf}                  & 74.43                       & 5.625                            & 0.0207                      & 0.305                     & 68.47                     & 16.023                    & 0.0442                      & 0.067                     & 10.0          \\
                                                            & GraphDF~\cite{luo2021graphdf}                  & 93.88                       & 10.928                           & 0.0636                      & 0.098                     & 90.61                     & 33.546                    & 0.1770                      & 0.000                     & 11.0          \\ \midrule
            \multirow{4}{*}{\rotatebox{90}{Diffusion}}      & EDP\mbox{-}GNN~\cite{niu2020permutation}       & 47.52                       & 2.680                            & 0.0046                      & 0.327                     & 87.21                     & 76.737                    & 0.0485                      & 0.000                     & 10.3          \\
                                                            & GDSS~\cite{jo2022score}                        & 95.72                       & 2.900                            & 0.0038                      & 0.698                     & 97.44                     & 14.656                    & 0.0021                      & 0.047                     & 8.0           \\
                                                            & DiGress~\cite{vignac2022digress}               & 98.19                       & \ul{0.095}                       & 0.0033                      & \ul{0.935}                & 94.99                     & 3.482                     & 0.0021                      & 0.416                     & 5.8           \\
                                                            & GruM~\cite{jo2023graph}                        & 99.69                       & 0.108                            & \textbf{0.0002}             & \textbf{0.945}            & \textbf{98.95}            & \ul{2.257}                & 0.0015                      & \textbf{0.530}            & 2.0           \\ \midrule
            \multirow{5}{*}{\rotatebox{90}{Inference-Time}} & Best\mbox{-}of\mbox{-}N~\cite{ma2025inference} & 99.22                       & 0.153                            & \ul{0.0003}                 & 0.912                     & \ul{98.84}                & 3.803                     & 0.0013                      & 0.485                     & 4.8           \\
                                                            & Beam Search~\cite{ma2025inference}             & \ul{99.81}                  & 0.123                            & 0.0004                      & 0.918                     & 98.12                     & 2.553                     & 0.0015                      & \ul{0.502}                & 3.5           \\
                                                            & TAGMol~\cite{dorna2024tagmol}                  & 99.63                       & 0.142                            & 0.0004                      & 0.918                     & 98.73                     & 2.647                     & \ul{0.0012}                 & 0.488                     & 3.8           \\
                                                            & SVDD~\cite{li2024derivative}                   & 99.41                       & 0.205                            & 0.0006                      & 0.905                     & 98.42                     & 2.706                     & 0.0018                      & 0.418                     & 5.9           \\ \cmidrule(l){2-11}
                                                            & \textbf{TreeDiff (Ours)}                       & \textbf{99.89}              & \textbf{0.091}                   & 0.0004                      & 0.932                     & \textbf{98.95}            & \textbf{2.135}            & \textbf{0.0011}             & 0.492                     & 1.8           \\
            \bottomrule
        \end{tabular}
        }
\end{table*}

\begin{table}[!t]
  \caption{Unconditional molecular graph generation results on 3D QM9 and Drugs regarding validity \& uniqueness (V\&U), atom-level stability (AS), and molecule-level stability (MS).}
  \label{tab:3d uncondition}
  \centering
  \resizebox{0.9\columnwidth}{!}{
    \begin{tabular}{clccc|c|c}
      \toprule
                                                      &                                                & \multicolumn{3}{c}{\bf QM9} & \multicolumn{1}{c}{\bf Drugs} &                                                                 \\ \cmidrule(lr){3-5}\cmidrule(lr){6-6}
                                                      & \textbf{Method}                                & \textbf{V\&U $\uparrow$}    & \textbf{AS $\uparrow$}        & \textbf{MS $\uparrow$} & \textbf{AS $\uparrow$} & \textbf{A.R.} \\ \midrule
      \multirow{2}{*}{\rotatebox{90}{Trad.}}          & ENF~\cite{garcia2021n}                         & 39.4                        & 85.0                          & 4.9                    & --                     & 16.7          \\
                                                      & G\mbox{-}SchNet~\cite{gebauer2019symmetry}     & 80.3                        & 95.7                          & 68.1                   & --                     & 15.3          \\ \midrule
      \multirow{10}{*}{\rotatebox{90}{Diffusion}}     & GDM~\cite{hoogeboom2022equivariant}            & --                          & 97.0                          & 63.2                   & 75.0                   & 15.3          \\
                                                      & GDM\mbox{-}Aug~\cite{hoogeboom2022equivariant} & 89.5                        & 97.6                          & 71.6                   & 77.7                   & 12.5          \\
                                                      & EDM~\cite{wu2022diffusion}                     & 90.7                        & \ul{98.7}                     & 82.6                   & 81.3                   & 8.0           \\
                                                      & EDM\mbox{-}Bridge~\cite{wu2022diffusion}       & 90.7                        & \textbf{98.8}                 & 84.6                   & 82.4                   & 6.8           \\
                                                      & GCDM~\cite{morehead2024geometry}               & 93.3                        & 98.7                          & 85.7                   & \textbf{89.0}          & 3.8           \\
                                                      & MiDi~\cite{vignac2023midi}                     & \ul{95.0}                   & 97.9                          & 84.0                   & 82.5                   & 7.0           \\
                                                      & GraphLDM~\cite{xu2023geometric}                & 82.7                        & 97.2                          & 70.5                   & 76.2                   & 14.0          \\
                                                      & GraphLDM\mbox{-}Aug~\cite{xu2023geometric}     & 89.5                        & 97.9                          & 78.7                   & 79.6                   & 11.2          \\
                                                      & GeoLDM~\cite{xu2023geometric}                  & 92.3                        & 98.2                          & 80.7                   & 79.3                   & 10.2          \\
                                                      & LDM\mbox{-}3DG~\cite{you2024latent}            & \textbf{95.3}               & 97.6                          & \ul{87.7}              & 80.51                  & 6.2           \\ \midrule
      \multirow{5}{*}{\rotatebox{90}{Inference-Time}} & Best-of-N~\cite{ma2025inference}               & 93.9                        & 98.6                          & 87.5                   & 85.4                   & 4.0           \\
                                                      & Beam Search~\cite{ma2025inference}             & 93.2                        & 98.5                          & 87.1                   & 84.6                   & 5.5           \\
                                                      & TAGMol~\cite{dorna2024tagmol}                  & 92.9                        & 98.7                          & 85.5                   & 85.1                   & 5.2           \\
                                                      & SVDD ~\cite{li2024derivative}                  & 92.8                        & 98.4                          & 86.3                   & 84.2                   & 6.8           \\ \cmidrule(l){2-7}
                                                      & \textbf{TreeDiff (Ours)}                       & 94.8                        & \textbf{98.8}                 & \textbf{88.6}          & \ul{86.5}              & 1.8           \\
      \bottomrule
    \end{tabular}}
\end{table}

\noindent\textbf{Conditional Generation on 3D Molecular Graphs.}
We further evaluate on the 3D QM9 benchmark~\citep{ramakrishnan2014quantum,hoogeboom2022equivariant}, where each molecule is annotated with six SE(3)-invariant quantum properties: polarizability $\alpha$ (Bohr$^3$), HOMO–LUMO gap $\Delta\varepsilon$ (meV), HOMO energy $\varepsilon_H$ (meV), LUMO energy $\varepsilon_L$ (meV), dipole moment $\mu$ (D), and heat capacity $C_v$ (cal/mol$\cdot$K). These properties are used as conditional targets, and we measure performance by the MAE between the requested conditions and oracle-predicted values. Two evaluation settings are considered: in-distribution (ID), where conditions are sampled within the training range, and out-of-distribution (OOD), where targets are shifted beyond it.
Baselines include a random sampler~\citep{you2024latent}, 3D diffusion models (EDM~\citep{wu2022diffusion}, LDM-3DG~\citep{you2024latent}), and inference-time guidance methods (Best-of-N~\citep{ma2025inference}, Beam Search~\citep{ma2025inference}, TAGMol~\citep{dorna2024tagmol}, and SVDD~\citep{li2024derivative}), all built on the LDM-3DG backbone under matched compute. On QM9 (Table~\ref{tab:3d condition}), TreeDiff achieves the best overall performance across both ID and OOD regimes, with the lowest average rank (A.R.\ 1.4). In ID, it reduces error on key properties such as polarizability and the HOMO–LUMO gap
, while also improving on $\varepsilon_L$, $\mu$, and $C_v$. Under distribution shift, the advantage becomes more pronounced and consistent across all six targets
. We attribute this robustness to long-horizon planning, which aligns intermediate denoising states with global property targets, and step-wise verification, which filters unstable edits and prevents error accumulation. Together, these mechanisms enable reliable conditional generation under both ID and OOD conditions.

\subsection{Unconditional Graph Generation}

\noindent\textbf{Unconditional Generation on 2D Molecular Graphs.} To assess model performance without explicit conditional supervision in training, we evaluate unconditional molecular graph generation on QM9~\citep{ramakrishnan2014quantum} and ZINC250k~\citep{irwin2012zinc}. Following standard practice, we report Validity (chemical sanity without rule-based repair), distributional fidelity via Fr\'echet ChemNet Distance (FCD)~\citep{preuer2018frechet}, global structure mismatch via the NSPDK kernel distance~\citep{costa2010fast}, and Scaffold similarity. These metrics also serve as guidance signals for inference-time baselines. We compare against traditional methods (MoFlow~\citep{zang2020moflow}, GraphAF~\citep{shi2020graphaf}, GraphDF~\citep{luo2021graphdf}), diffusion-based generators (EDP-GNN~\citep{niu2020permutation}, GDSS~\citep{jo2022score}, DiGress~\citep{vignac2022digress}, GruM~\citep{jo2023graph}), and inference-time guidance strategies (Best-of-$N$~\citep{ma2025inference}, Beam Search~\citep{ma2025inference}, TAGMol~\citep{dorna2024tagmol}, SVDD~\citep{li2024derivative}). We also report the training set as a reference upper bound for distributional metrics. As shown in Table~\ref{tab:2d uncondition}, TreeDiff achieves the lowest FCD on both datasets and near-perfect validity, while remaining competitive on NSPDK and scaffold similarity. Compared with both diffusion backbones and inference-time guidance methods, TreeDiff provides a more reliable balance between distributional fidelity and chemical plausibility. These results highlight the effectiveness of planning-guided sampling in advancing unconditional molecular graph generation.

\noindent\textbf{Unconditional Generation on 3D Molecular Graphs.} We further evaluate TreeDiff on unconditional 3D molecular generation using QM9~\citep{ramakrishnan2014quantum} and Drugs~\citep{axelrod2022geom} as benchmarks. Following established protocols~\citep{hoogeboom2022equivariant,wu2022diffusion,vignac2023midi,morehead2024geometry}, we report validity \& uniqueness (V\&U), atom-level stability (AS), and molecule-level stability (MS) on QM9, and AS on Drugs. Baselines span a broad range of 3D generators, including traditional methods (ENF~\citep{garcia2021n}, G-SchNet~\citep{gebauer2019symmetry}); diffusion-based models (GDM~\citep{hoogeboom2022equivariant}, GDM-Aug, EDM~\citep{wu2022diffusion}, EDM-Bridge, GCDM~\citep{morehead2024geometry}, MiDi~\citep{vignac2023midi}, GraphLDM~\citep{xu2023geometric}, GraphLDM-Aug, GeoLDM, LDM-3DG~\citep{you2024latent}); and inference-time strategies (Best-of-$N$~\citep{ma2025inference}, Beam Search, TAGMol~\citep{dorna2024tagmol}, SVDD~\citep{li2024derivative}). As shown in Table~\ref{tab:3d uncondition}, TreeDiff achieves the best overall performance on QM9 and the second-best AS on Drugs, yielding the lowest overall average rank (A.R.\ 1.8) across benchmarks. On QM9, it attains the highest molecule-level stability (MS 88.6) and state-of-the-art atom stability tied with EDM-Bridge (AS 98.8) while maintaining strong validity and uniqueness (94.8). Compared with diffusion backbones (e.g., LDM\mbox{-}3DG, GCDM), TreeDiff substantially reduces whole-molecule failures through global planning. Against inference-time methods, it preserves their high atom-level reliability but produces more globally consistent structures. On Drugs, TreeDiff maintains strong atom stability (86.5), confirming its robustness on larger, drug-like molecules. Overall, these results highlight TreeDiff's ability to couple local atomic precision with global structural coherence in 3D molecular generation.

\subsection{Ablation Study}

\begin{table}[!t]
    \caption{Ablation results of the dual-space denoising on 2D/3D unconditional generation tasks, where $\sigma$ is used to control the impact of the dual-space denoising, where a large value indicate less impact. }
    \label{tab:ablation denoising}
    \centering
    \resizebox{\columnwidth}{!}{
        \begin{tabular}{lccc|ccc}
            \toprule
                                             & \multicolumn{3}{c}{\textbf{2D QM9}} & \multicolumn{3}{c}{\textbf{3D QM9}}                                                                                                            \\ \cmidrule(lr){2-4}\cmidrule(lr){5-7}
            Variant                          & \textbf{Valid $\uparrow$}           & \textbf{FCD $\downarrow$}           & \textbf{NSPDK $\downarrow$} & \textbf{V\&U $\uparrow$} & \textbf{AS $\uparrow$} & \textbf{MS $\uparrow$} \\ \midrule
            TreeDiff ($\mathbf{\sigma=1.0}$) & \textbf{99.90}                      & \textbf{0.090}                      & \textbf{0.00040}            & \textbf{94.8}            & \textbf{98.8}          & \textbf{88.6}          \\ \midrule
            $\mathbf{\sigma=0.1}$            & 97.56                               & 0.256                               & 0.00100                     & 92.5                     & 98.6                   & 86.9                   \\
            $\mathbf{\sigma=0.5}$            & 99.48                               & 0.112                               & 0.00055                     & 93.4                     & 98.7                   & 87.4                   \\
            $\mathbf{\sigma=10}$             & 99.15                               & 0.125                               & 0.00064                     & 92.4                     & 98.5                   & 86.8                   \\ \midrule
            w/o graph space                   & 98.13                               & 0.156                               & 0.00087                     & 92.1                     & 97.2                   & 86.8                   \\
            \bottomrule
        \end{tabular}}
\end{table}

\begin{table}[!t]
    \caption{Ablation results on the verifier on 2D/3D conditional generation, where Latent + GPM is the default. }
    \label{tab:ablation verifier}
    \centering
    \resizebox{\columnwidth}{!}{
        \begin{tabular}{lcc|cc|cc}
            \toprule
                                                         & \multicolumn{2}{c}{\textbf{BBBP}} & \multicolumn{2}{c}{\textbf{QM9 (ID)}} & \multicolumn{2}{c}{\textbf{QM9 (OOD)}}                                                                                                                  \\ \cmidrule(lr){2-3}\cmidrule(lr){4-5}\cmidrule(lr){6-7}
            \textbf{Verifier Type}                       & \textbf{Acc $\uparrow$}           & \textbf{MAE $\downarrow$}             & $\alpha$ \textbf{$\downarrow$}         & $\varepsilon_H$ \textbf{$\downarrow$} & $\alpha$ \textbf{$\downarrow$} & $\varepsilon_H$ \textbf{$\downarrow$} \\ \midrule
            Latent-only                                  & 92.4                              & 0.405                                 & 17.2                                   & 69.5                                  & 36.0                           & 116.2                                 \\ \midrule
            Graph-only (GIN~\citep{xu2018how})           & 92.8                              & 0.395                                 & 16.6                                   & 68.3                                  & 34.8                           & 113.4                                 \\
            Graph-only (GT~\citep{rampavsek2022recipe})  & 93.3                              & 0.380                                 & 15.8                                   & 64.5                                  & 32.5                           & 106.5                                 \\
            Graph-only (GPM~\citep{wang2025beyond})      & 93.8                              & 0.381                                 & 15.6                                   & 65.3                                  & 33.6                           & 104.1                                 \\ \midrule
            Latent + GIN~\citep{xu2018how}               & 93.8                              & 0.374                                 & 16.1                                   & 66.8                                  & 33.5                           & 110.7                                 \\
            Latent + GT~\citep{rampavsek2022recipe}      & 94.5                              & 0.350                                 & 15.3                                   & 63.0                                  & 31.0                           & 103.5                                 \\
            \textbf{Latent + GPM~\citep{wang2025beyond}} & \textbf{94.8}                     & \textbf{0.342}                        & \textbf{14.9}                          & \textbf{61.6}                         & \textbf{28.6}                  & \textbf{99.5}                         \\
            \bottomrule
        \end{tabular}}
\end{table}

\noindent\textbf{Ablation on dual-space denoising.}
We investigate the impact of dual-space denoising strength by varying the guidance scale $\sigma\!\in\!\{0.1,0.5,1.0,10\}$ on 2D and 3D QM9 datasets under unconditional settings. A \emph{w/o graph space} variant is also included as a latent diffusion-only baseline. As shown in Table~\ref{tab:ablation denoising}, excessively strong guidance ($\sigma{=}0.1$) over-constrains the denoising trajectory, leading to degraded validity and fidelity, whereas overly weak guidance ($\sigma{=}10$) approaches standard diffusion and offers limited improvement. The default configuration ($\sigma{=}1.0$) achieves the best overall performance across both 2D and 3D metrics, demonstrating that moderate dual-space coupling substantially enhances generation quality compared to the diffusion-only baseline.

\noindent\textbf{Ablation on the verifier.}
We examine the impact of verifier design under identical computational budgets on conditional generation tasks across BBBP, QM9 (ID), and QM9 (OOD). We consider three categories of verifiers: \emph{Latent-only}, \emph{Graph-only} (GIN, Graph Transformer (GT), and GPM), and \emph{Latent+Graph} (Latent+GIN, Latent+GT, Latent+GPM). As summarized in Table~\ref{tab:ablation verifier}, the Latent-only verifier provides fast feedback but lacks discrete structural cues, resulting in weaker calibration. Graph-only variants better capture structural constraints: GIN leverages local substructure sensitivity, while GT reduces prediction errors through long-range aggregation. Combining latent and graph modalities consistently improves performance—both \emph{Latent+GIN} and \emph{Latent+GT} surpass their single-modality counterparts. Our default \emph{Latent+GPM} achieves the strongest results overall, indicating that global attention enhanced by graph-structural priors yields more accurate and stable node-wise calibration across noise scales.

\subsection{Efficiency Analysis}

\begin{table}[!t]
    \caption{Efficiency analysis on the verifier for 2D/3D conditional generation. }
    \label{tab:efficiency verifier}
    \centering
    \resizebox{\columnwidth}{!}{
        \begin{tabular}{lccc|ccc}
            \toprule
                                                 & \multicolumn{3}{c}{\textbf{BACE}} & \multicolumn{3}{c}{\textbf{QM9 (ID)}}                                                                                                         \\            \cmidrule(lr){2-4}\cmidrule(lr){5-7}
            \textbf{Variant}                     & \textbf{Acc $\uparrow$}           & \textbf{MAE $\downarrow$}             & \textbf{Comp.} & $\varepsilon_L$ \textbf{$\downarrow$} & $C_v$ \textbf{$\downarrow$} & \textbf{Comp.} \\ \midrule
            Post-hoc + TreeDiff Verify           & 87.1                              & 0.409                                 & 9.6x           & 59.5                                  & 13.0                        & 15.6x          \\
            Post-hoc + Oracle Verify             & 86.7                              & 0.420                                 & 8.9x           & 60.0                                  & 13.5                        & 14.2x          \\ \midrule
            \textbf{Step-wise + TreeDiff Verify} & 87.8                              & 0.392                                 & 1.0x           & 60.2                                  & 13.4                        & 1.0x           \\
            \bottomrule
        \end{tabular}}
\end{table}

\noindent\textbf{Efficiency analysis on the verifier.}
We compare \emph{Post-hoc} verification, which evaluates only the final denoising step ($t{=}0$), against two baselines—\emph{Post-hoc + Oracle Verify} and \emph{Post-hoc + TreeDiff Verify}—as well as our proposed \emph{Step-wise + TreeDiff Verify}. As shown in Table~\ref{tab:efficiency verifier}, step-wise verification achieves the best overall balance between accuracy and efficiency, reducing computation cost by over $9\times$ on BACE and $15\times$ on QM9 (ID) while maintaining comparable or even slightly better predictive accuracy. In contrast, post-hoc variants require substantially higher compute since verification occurs only after full trajectory generation.

\begin{figure}[!t]
    \centering
    \includegraphics[width=\linewidth]{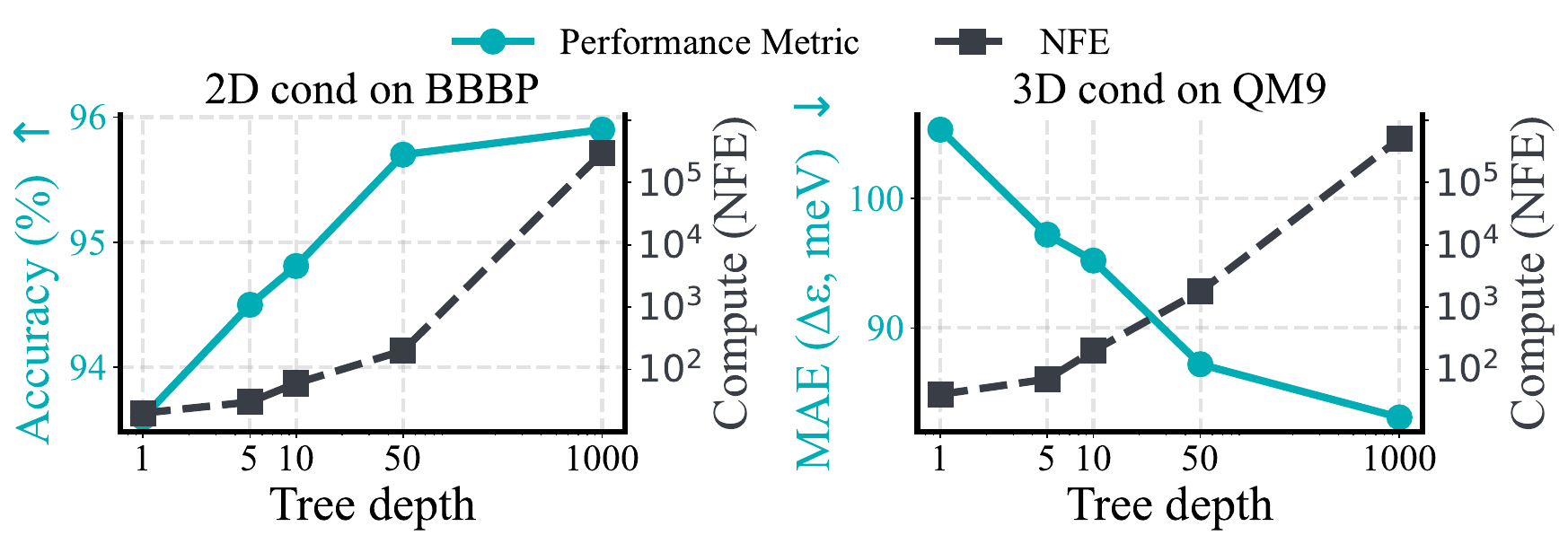}
    \caption{Efficiency analysis on the MCTS tree depth vs.\ inference-time computation.}
    \label{fig:efficiency tree depth}
\end{figure}

\noindent\textbf{Efficiency analysis on tree depth.}
We vary the macro-step budget (tree depth) $D\!\in\!\{1,5,10,50,1000\}$ and report task performance alongside inference compute (NFE) for two conditional generation settings: 2D BBBP and 3D QM9 (ID). As reported in Figure \ref{fig:efficiency tree depth}, model performance improves rapidly for small depths between $D{=}1$ and $D{=}10$ then plateaus beyond $D{\ge}50$. Meanwhile, NFE grows near-geometrically with $D$ due to repeated expansions and evaluations, revealing a clear compute-quality trade-off. A practical Pareto regime emerges at $D{\in}[10,50]$, which captures most quality gains while keeping computation moderate.
\section{Conclusion}

We propose TreeDiff, an inference-time framework that couples MCTS with dual-space diffusion to enable controllable and stable graph generation without retraining. The architecture comprises three key components: (i) \emph{macro-step expansion} that operates over groups of denoising steps to reduce search depth, (ii) \emph{dual-space denoising} that jointly leverages latent efficiency and graph-level structural fidelity, and (iii) a \emph{dual-space verifier} that provides rollout-free node-wise evaluation for early pruning. Extensive experiments on 2D/3D molecular generation, across unconditional and conditional regimes and ID/OOD settings, demonstrate the state-of-the-art performance and favorable inference-time scaling of TreeDiff.

\begin{acks}
The work of Z. Wang, C. Zhang, and Y. Ye was partially supported by the NSF under grants IIS-2533550, IIS-2321504, IIS-2217239, CNS-2426514, and CMMI-2146076, Notre Dame Strategic Framework Research Grant (2025), and Notre Dame Poverty Research Package (2025). Any expressed opinions, findings, and conclusions or recommendations are those of the authors and do not necessarily reflect the views of the sponsors.
\end{acks}

\bibliographystyle{ACM-Reference-Format}
\bibliography{references}

\balance
\appendix

\section{Pseudo Code}
\label{sec:inference}
TreeDiff’s inference procedure is summarized in Algorithm~\ref{alg:treediff}.

\begin{algorithm}[!h]
       \caption{TreeDiff Inference}
       \label{alg:treediff}
       \begin{algorithmic}[1]
              \Require Diffusion predictor $\epsilon_\theta$; total steps $T$; start step $t_s$;
              encoder/decoder $(\mathrm{Enc},\mathrm{Dec})$; discrete refiner $p_\psi$;
              verifier $V_\phi$; max children $K$; rollouts per round $N_r$;
              depth budget $D_{\max}$; UCB constant $c_{\mathrm{ucb}}$.
              \Ensure Final graph $\mathcal{G}_0$.

              \State \textbf{Initialize:} latent root $z_{t_s}$; set node $\mathcal{R}\!\leftarrow\!z_{t_s}$ with stats $Q{=}0$, $N{=}0$;
              set remaining depth $D{\leftarrow}D_{\max}$.

              \While{$t_s>0$}
              \For{$i=1$ to $N_r$}
              \State \textbf{Selection}: From node $u$ (latent state $z_t$), descend by UCB until a leaf/terminal.
              \State \textbf{Expansion}: At node $u$ with time $t$, propose up to $K$ children via:
              \begin{itemize}
                     \item \emph{Jump (macro-step)}: sample a step length $k$ (e.g., around $\lfloor t/D \rfloor$) and advance $z_t \!\to\! z_{t-k}$ by $k$ reverse latent updates.
                     \item \emph{Dual-Space Denoising (Sec.\ Dual-Space Denoising)}:
                           (a) decode $\mathcal{G}_{t-k}^{\text{str}} \!=\! \mathrm{Dec}(z_{t-k},\, t{-}k)$;
                           (b) apply lightweight discrete refinement with $p_\psi$;
                           (c) re-encode to obtain a structure-aware latent $\tilde z_{t-k}$ to anchor the child state.
              \end{itemize}
              \Statex \hspace{1.68em}The resulting child stores both latent $\tilde z_{t-k}$ and its decoded graph $\tilde{\mathcal{G}}_{t-k}^{\text{str}}$.
              \State \textbf{Simulation} (Sec.\ Dual-Space Verifier): For each child, score with
              $V_\phi\big(\tilde z_{t-k},\, \tilde{\mathcal{G}}_{t-k}^{\text{str}},\, t{-}k\big)$;
              initialize $Q(\cdot)$ and $N(\cdot)$ accordingly.
              \State \textbf{Backpropagation}: Propagate the child score along the selected path (update visits and values).
              \EndFor
              \State \textbf{Commit}: Select the best child of the root by average value; set it as new root $\mathcal{R}$.
              Update $t_s \!\leftarrow\! t_s - k(\mathcal{R})$, $D \!\leftarrow\! \max(D{-}1,1)$.
              (Optionally prune siblings to keep top-$M$.)
              \EndWhile

              \State \textbf{Return}: $\mathcal{G}_0 \leftarrow \mathrm{Dec}(z_0,\, 0)$.
       \end{algorithmic}
\end{algorithm}

\noindent\textbf{Procedure.}
Starting from a latent root $z_{t_s}$, TreeDiff performs a sequence of short MCTS rounds to construct a single high-quality denoising path. In each round, \textbf{Selection} first descends from the current root to a leaf via UCB, focusing computation on promising branches. Given the selected leaf at time $t$, \textbf{Expansion} then proposes up to $K$ children in two phases: a macro-step jump that advances $k$ reverse updates in latent space, followed by \emph{dual-space denoising}—decode to graph space, apply lightweight discrete refinement, and re-encode—to anchor a structure-aware latent state. Next, \textbf{Simulation} evaluates each child using the dual-space verifier on both the latent and its decoded graph, avoiding costly full rollouts while preserving structural feedback. Finally, \textbf{Backpropagation} updates visit counts and value estimates along the traversed path. After accumulating $N_r$ rollouts, TreeDiff \emph{commits} the best child as the new root, reduces $t$ by $k$, and repeats until $t{=}0$; the resulting $z_0$ is then decoded to $\mathcal{G}_0$.

\section{Time Complexity}
Let $C_{\text{lat}}$ be the cost of one latent reverse step, $C_{\text{ref}}$ the cost of one dual-space refinement
(decode $+$ discrete refine $+$ re-encode), and $C_{\text{ver}}$ the cost of one verifier call.
At each visited node, UCB-based \textbf{Selection} scans its children; if the branching factor is bounded by $K$,
the path cost is $O(K\,D_{\max})$ (or $O(D_{\max})$ when child stats are cached and $K$ is treated as a constant).

For a leaf that is expanded into up to $K$ children, the per-child cost of
\textbf{Expansion}$+$\textbf{Simulation} is
\[
       O\!\big(k\,C_{\text{lat}} + C_{\text{ref}} + C_{\text{ver}}\big),
\]
where $k$ is the sampled macro-step length for that child.
Therefore, one MCTS \emph{round} with $N_r$ rollouts incurs
\[
       O\!\Big(
       N_r\big[ K\,D_{\max} \;+\; K\big(k\,C_{\text{lat}} + C_{\text{ref}} + C_{\text{ver}}\big) \big]
       \Big).
\]
Across the whole inference, the number of commits is roughly $T/\bar{k}$ (with $\bar{k}$ the average macro-step), and
never exceeds $D_{\max}$. Using $\bar{k}$ to summarize $k$’s variability, the total work is
\[
       O\!\Big(
       \frac{T}{\bar{k}}\; N_r\big[ K\,D_{\max} \;+\; K\big(\bar{k}\,C_{\text{lat}} + C_{\text{ref}} + C_{\text{ver}}\big) \big]
       \Big).
\]
If $K$ is constant and selection bookkeeping is $O(1)$ per level,
this simplifies to
\[
       O\!\Big(
       \frac{T}{\bar{k}}\; N_r \big[ D_{\max} \;+\; \bar{k}\,C_{\text{lat}} + C_{\text{ref}} + C_{\text{ver}} \big]
       \Big).
\]
\noindent\textbf{Discussion.}(i) \textbf{Trade-off via $\bar{k}$.} Larger $\bar{k}$ reduces the number of commit steps ($T/\bar{k}$)
but increases the per-child expansion cost ($\bar{k}C_{\text{lat}}$); small $\bar{k}$ does the opposite.
(ii) \textbf{Verifier amortization.} When $V_\phi$ is a lightweight MLP,
$C_{\text{ver}} \ll \bar{k}C_{\text{lat}}$, so runtime is dominated by latent steps and occasional refinement.
If $V_\phi$ batches $B$ children on accelerator, its \emph{amortized} cost becomes $C_{\text{ver}}/B$.
(iii) \textbf{Parallelism.} The $K$ children of a leaf and the $N_r$ rollouts within a round are embarrassingly parallel;
with $P$ workers the effective factor becomes $K/P$ or $N_r/P$ in the bracketed term.
(iv) \textbf{Degenerate cases.} With $K{=}1$ and $N_r{=}1$, TreeDiff reduces to a single guided trajectory with cost
$O\!\big(T\,C_{\text{lat}} + \tfrac{T}{\bar{k}}(C_{\text{ref}}{+}C_{\text{ver}})\big)$.
If dual-space refinement is skipped ($C_{\text{ref}}{=}0$), complexity coincides with pure latent MCTS.
(v) \textbf{Memory.} Active-node storage is $O(N_r K)$ per round without pruning, or $O(M D_{\max})$
when keeping only top-$M$ siblings per level.

\section{Implementation Details}
\label{sec:imp}

All diffusion backbones are pretrained under standard noise schedules before integration into TreeDiff. For macro-step expansion, we set the tree depth to $D{=}10$ and expansion width to $B{=}4$, with an adaptive expansion variance $\sigma_k{=}0.1$, and use PUCT as the selection policy. For dual-space denoising, the guidance scale is $\sigma_g{=}1.0$ and $h_{t'}{=}1.0$; the latent denoising horizon is set to half of the total diffusion steps within each macro step ($n{=}k/2$), while the structural refinement length is set to 10\% of that value ($m{=}k/10$). The encoder, decoder, and verifier are trained through trajectory distillation and remain frozen during search. For 2D molecular generation, we employ a Graph Transformer~\citep{rampavsek2022recipe} as the encoder–decoder backbone, and for 3D molecular generation, we use a SE(3)-Transformer~\citep{fuchs2020se}. The verifier adopts the recently proposed GPM~\citep{wang2025beyond}, which excels at capturing substructural dependencies in graphs.

\end{document}